\documentclass{article}
\PassOptionsToPackage{numbers, compress}{natbib}

\usepackage[dblblindworkshop, final]{neurips_2026}
\workshoptitle{AI for Science and SIMBIOCHEM workshops (NeurIPS 2026)}

\usepackage[utf8]{inputenc} 
\usepackage[T1]{fontenc}    
\usepackage{hyperref}       
\usepackage{url}            
\usepackage{booktabs}       
\usepackage{amsfonts}       
\usepackage{nicefrac}       
\usepackage{microtype}      
\usepackage{xcolor}         
\usepackage{graphicx} 
\usepackage{wrapfig}
\usepackage{amsmath}

\title{From Simulated Fluids to Living Tissue: Zero-Shot Foundation Model Transfer to Perturbed Embryos}

\author{%
  Louisa Cornelis \\
  UC Santa Barbara \\
  PolymathicAI \\
  \And
  Haewon Jeong \\
  UC Santa Barbara \\
  PolymathicAI \\
  \AND
  Irina Espejo\thanks{Co-last authors} \\
  New York University \\
  PolymathicAI \\
  \AND
  Payel Mukhopadhyay\footnotemark[1] \\
  University of Cambridge, UK \\
  PolymathicAI
}

\begin{document}
\maketitle

\begin{abstract}

Morphogenesis is driven by the interplay of genetic regulation and tissue mechanics. \textit{In vivo} video data is limited, particularly under genetic mutations and environmental perturbations, making generalization beyond the most representative conditions in the dataset challenging. However, at tissue scales, the dynamics of \textit{Drosophila} can be described as a continuum flow, motivating the usage of representations learned across other physical systems as priors for developmental dynamics.

We investigate finetuning Walrus, a Continuum Dynamics Foundation Model pretrained across a broad set of simulated continuous systems, on tissue velocity fields from only 12 wild-type \textit{Drosophila} blastoderm videos. We evaluate autoregressive prediction both in-distribution and zero-shot under unseen temperature and genetic perturbations. Zero-shot refers to evaluation on perturbation conditions never seen during wild-type finetuning, without any additional training. In-distribution, a Fourier Neural Operator model trained from scratch achieves best performance on VRMSE. However, Walrus (CRPS) is the strongest model under the most severe distribution shift of unseen samples of mutated embryos and temperature conditions. Interestingly, predictions capture the shape of the mean surface tissue velocity trajectory and its separation between $17^\circ$C and $27^\circ$C, including the timing of the peak at Germ Band Extension onset, a biologically meaningful milestone of embryo development.

Our results provide preliminary support that representations learned from different dynamical systems can transfer to experimental morphogenesis in developmental regimes, providing they admit a continuum description. More broadly, our results motivate further study of physics foundation models as a new complementary tool for the simulation of morphogenesis in low-data regimes.

\end{abstract}

\section{Introduction}

Understanding morphogenesis, the process by which a cell, tissue, or organism acquires its shape, remains one of modern biology's central challenges. Shape emerges as cells multiply, grow, deform, and migrate. Over just a few hours, an unpatterned sheet can specialize into an organized body plan with distinct segments defining the head, thorax, and abdomen. The mechanisms orchestrating this transformation are only partly understood, with consensus converging towards a complex interplay between gene expression and physical mechanics at different scales \citep{heisenberg2013forces, collinet2021programmed}. Beyond fundamental biology, understanding structural emergence is an essential piece to the development of interventions on embryonic development, tissue repair \citep{wood2002wound}, and cancer progression \citep{friedl2009collective}. These processes have been extensively studied in model organisms such as the zebrafish and the fruit fly (\emph{Drosophila}).

Modeling the dynamics of morphogenesis has traditionally implied focusing on two paradigms. On one hand, continuum descriptions model tissue as an active material governed by PDEs, naturally scaling to embryo-wide fields \citep{streichan2018global, alber2025model}. These equations can be derived by hand or recovered directly from data by sparse regression \citep{brunton2016discovering, rudy2017data}. On the other hand, discrete cell-based models take the opposite approach and resolve the mechanics of individual cells \citep{honda2022mathematical, fletcher2014vertex}. Here, the modeler prescribes the rules and parameters that every cell obeys, and simulating large populations over long trajectories remains computationally demanding \citep{fletcher2022seven}. Ultimately, whether using these mechanistic paradigms or specialized machine learning models \citep{yang2026multicell, lefebvre2024learning, huang2025learning}, the result is fit exclusively to the target system. This shared reliance on single-domain data hinders generalization across stages, genotypes, or organisms, meaning every new problem requires starting from scratch.

Foundation models offer an alternative that directly addresses this generalization bottleneck. By pretraining a single backbone across diverse, heterogeneous physical systems, these models learn robust dynamic representations that can transfer to entirely new domains \citep{mccabe2024multiple, hao2024dpot, herde2024poseidon}. In this paper, we ask whether a model that has learned continuum dynamics elsewhere can adapt to the morphogenesis setting. We finetune Walrus, a transformer-based Foundation Model for continuum dynamics \citep{mccabe2026walrus}, on live imaging of \textit{Drosophila} blastoderm development. To our knowledge, this represents the first time a physics foundation model has been transferred to morphogenesis.

Using Walrus is physically motivated. In developing tissues, cellular structures behave as continuous, liquid-like active materials that exhibit laminar flow-like behavior, acting as viscoelastic solids in the short term and flowing like viscous fluids over longer developmental timescales \citep{forgacs1998viscoelastic, foty1996surface, foty1994liquid}. Because Walrus is pretrained on nineteen continuum systems that closely mirror these mechanical regimes, such as viscoelastic instability, incompressible shear flow, Gray-Scott reaction-diffusion, and active matter \citep{ohana2024well, tali2024flowbench}, we posit that it can provide two distinct advantages over models trained from scratch. First, it should enable data-efficient transfer, which is critical given that \textit{in toto} (whole-embryo) light-sheet imaging yields at most hundreds of embryos. Second, it should provide superior out-of-distribution robustness. Walrus has previously demonstrated this efficiency by recovering experimental dynamics of the Rayleigh-Taylor instability from as few as one to three simulations of a previously unseen system \citep{mukhopadhyay2026emergent}. Here, we test whether that transfer capacity extends to the living tissue of an embryo.

\begin{figure}[t]
  \centering
  \includegraphics[width=\linewidth]{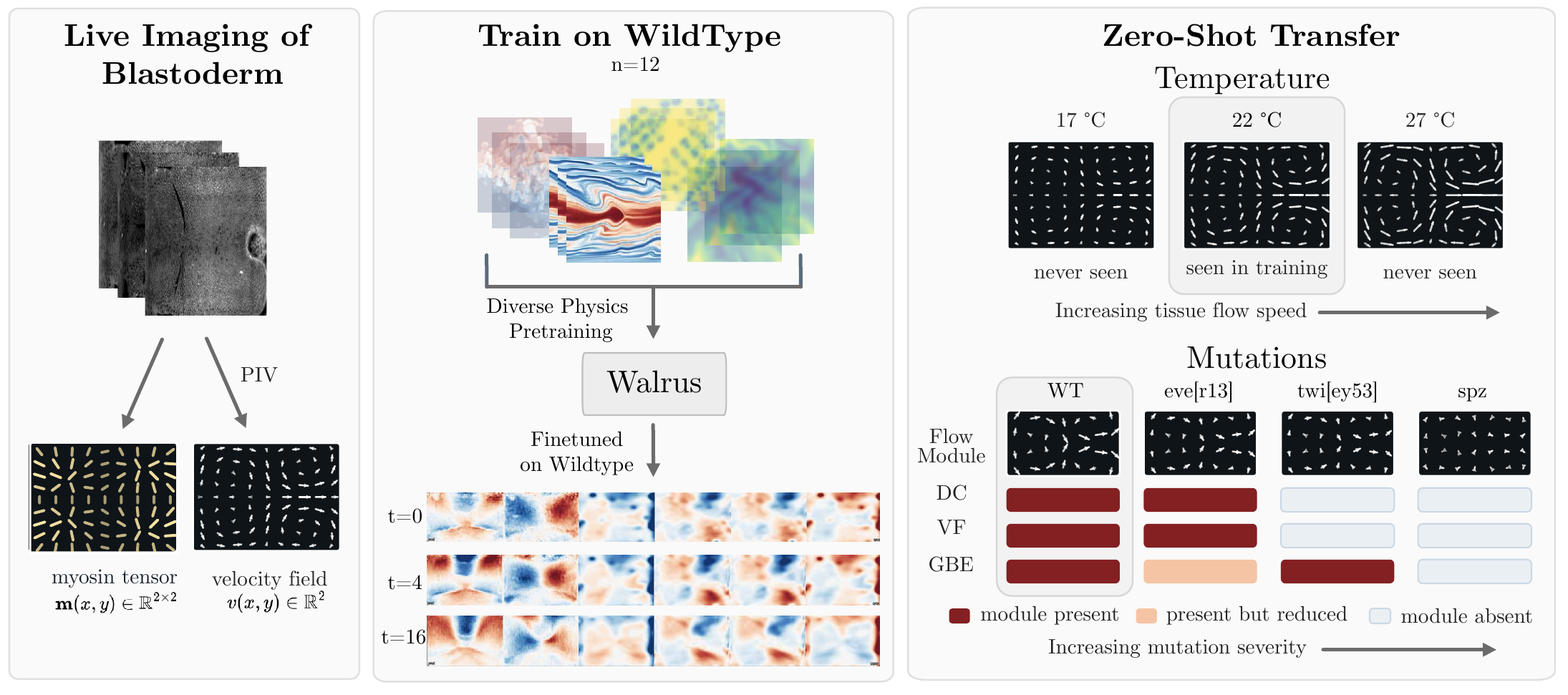}
  \caption{\textbf{Overview of the Walrus finetuning workflow and zero-shot experiments.} Experimental videos of live blastoderm are converted to velocity-field representations (left) and used to finetune Walrus on \textit{only} wild-type embryo videos (middle). The finetuned model is evaluated on zero-shot prediction on unseen deviations of wild-type development (right): first on temperatures 17 and 27 $^\circ\text{C}$ (top) and second on three genotype variations (differing in dorsal contraction (DC), ventral furrow (VF), and germ band extension (GBE) modules) (bottom).}
  \label{fig:graphical-abstract}
\end{figure}

Specifically, we use \textit{Drosophila} blastoderm development due to the extensive literature and high-quality experimental data available \citep{jaeger2012drosophila, st1992origin}. We utilize the DynamicAtlas \citep{lefebvre2026dynamicatlas}, an established dataset providing \textit{in toto} videos across wild-type and 18 mutant genotypes. By successfully finetuning Walrus on wild-type embryos, we establish the first transfer of a physics foundation model to the morphogenesis setting, suggesting that representations of synthetic continuum dynamics can adapt to living tissue. Compared with strong domain baselines, including Fourier-based operator models (FFNO \citep{tran2023factorized}), vision-based architectures (Sine-Net \citep{zhang2024sinenet}), and other foundation models (Poseidon-L \citep{herde2024poseidon}), we demonstrate the advantages of our proposed method in the following ways:

\begin{itemize}
    \item \textbf{Predicting temperature-dependent timescale scaling:} In open-loop rollout settings, our fine-tuned model captures the qualitative trend of how shifts to 17$^\circ$C and 27$^\circ$C alter tissue flow speeds without prior thermal training.
    
    \item \textbf{Robustness across genetic mutation phenotypes:} The finetuned model successfully predicts high-level morphodynamic phenotypical changes ranging from reduced flow speeds in \textit{eve}[r13] to the complete absence of morphodynamic modules in severe knockouts like \textit{twi}[ey53] and \textit{spz}.

    \item \textbf{Outperforming established architectures:} The fine-tuned Walrus achieves lower or comparable VRMSE scores across most zero-shot tasks compared to the neural operator, convolutional, and foundation model baselines.
\end{itemize}

\section{Methods}

\subsection{Data Processing}

\textbf{Tissue Velocity Fields.}
Tissue flow fields were derived from DynamicAtlas cylindrical pullbacks of \textit{Drosophila} embryos expressing Sqh--mCherry or Sqh--GFP \citep{lefebvre2026dynamicatlas}. Of 20 recordings, 18 passed quality control assessments for unwrapping artifacts, luminosity fluctuations, and whole-field drift. See Table \ref{tab:t0} for the number of time frames available per embryo. We generated high-resolution velocity fields for all retained recordings using Particle Image Velocimetry (PIV) software, PIVlab \citep{thielicke2014pivlab}, to ensure consistency across embryos. The resulting fields were filtered to remove outliers, utilizing gap-filled components to reconstruct missing vectors. Extended quality control metrics and exclusion criteria are detailed in Appendix \ref{app:qc}.

\textbf{Grid and Walrus Formatting.} 
For model ingestion, the PIVlab spatial grids were mapped to a $64 \times 96$ resolution. To handle projection distortions at the anterior and posterior poles, we cropped the ends of the image and linearly resampled the remaining middle section. To fix uneven spacing at the seam and ensure the data still wraps continuously, we resampled the circumference to 96 evenly spaced points. Each embryo was written as a Well HDF5 trajectory \citep{ohana2024well}.

\textbf{Temporal Alignment.}
Embryos were aligned to a common developmental clock using the kinematic landmark $t_0^\omega$. This landmark is derived from the local tissue vorticity pattern of the surface velocity \citep{lefebvre2026dynamicatlas}. The alignment scores from our high-resolution PIV fields sometimes exhibit multiple peaks, so we modified the standard global-maximum alignment rule to instead use the first prominent local maximum of the vorticity autocorrelation score. This ensured our temporal alignment values more closely matched the reference landmarks established in the original paper. The exact formulation and alignment validations are provided in Appendix \ref{app:temporal}.

\textbf{Myosin Tensors.}
Myosin anisotropy tensors were extracted from the Sqh--mCherry fluorescence signal using a local windowed-Radon transform approach \citep{lefebvre2023geometric}. We extracted symmetric, traceless tensors from the raw maximum-projection intensities. Validation against reference atlas tensors demonstrated that utilizing raw maximum-projection intensities 
best preserved the original concentration-weighted tensor orientations and spatial norm structures, see Appendix \ref{app:myosin}.

\subsection{Models and Training}

\paragraph{Training Setup.}
All models share an identical data pipeline and trainer to isolate architecture. Models predict the next frame from a context window of past frames on a $64 \times 96$ embryo grid and roll out autoregressively during evaluation. 

We train on wild-type data (12 train, 3 validation, 3 test embryos) under two input settings: velocity only (2 tangent-velocity channels, $v_{AP}$ along the anteroposterior axis and $v_{DV}$ along the periodic dorsoventral/circumferential axis) and velocity with the myosin tensor (6 channels total: the two velocity channels plus the four stored tensor components $M_{AP,AP}$, $M_{AP,DV}$, $M_{DV,AP}$, and $M_{DV,DV}$). Inputs are normalized using global reversible instance normalization, and boundary conditions are passed via auxiliary tokens. Unless noted, models predict residuals to the next state, rather than full field values.

Training uses an $L_1$ loss (MAE) with AdamW~\citep{loshchilov2017decoupled} ($\text{learning rate} = 10^{-4}$, weight decay $= 10^{-4}$), a reciprocal square-root schedule~\citep{zhai2022scaling,hagele2024cooldown} (10-epoch warm-up from $0.1\times$, 10-epoch cooldown), gradient clipping at 10, batch size 1, and full \texttt{fp32} precision on a single H100 GPU. Following \textsc{Walrus}'s fixed-budget protocol, all models train for 100 epochs at 200 sampled windows per epoch ($2 \times 10^{4}$ total samples).

\textbf{Models and Baselines.}
To ensure a comprehensive evaluation, our selected baselines provide thorough coverage across major machine learning domains, spanning classical physics methods, convolutional networks, neural operators, and foundation models. See Table~\ref{tab:models} for a summary of each model's configuration.

\noindent \textbf{\textsc{Walrus}}~\citep{mccabe2026walrus}: 1.3B-parameter model finetuned with learnable rotary embeddings, full spatial attention, 40 processor blocks (width 1408), causal time attention, and patch jitter.

\textbf{\textsc{Walrus (scratch)}}~\citep{mccabe2026walrus}: The Walrus architecture trained entirely from scratch rather than finetuned from a pretrained checkpoint.

\noindent\textbf{\textsc{Walrus}\,+\,CRPS}~\citep{diaconu2026probabilistic}: A probabilistic variant finetuned from the best deterministic checkpoint. A 32D latent noise vector is injected via adaptive layer normalization into 20 of 40 blocks. Trained via a 4-member fair-CRPS estimator~\citep{alet2025skillful} (AdaLN $\text{lr} = 10^{-4}$, backbone $\text{lr} = 5 \times 10^{-5}$). Evaluation averages 4 sampled members for point metrics (VRMSE) and scores the ensemble for CRPS. This stage runs for an additional 100 epochs on top of the deterministic checkpoint’s 100.

\noindent \textbf{Poseidon-L}~\citep{herde2024poseidon}: A 629M-parameter shifted-window operator pretrained on PDEGym. Finetuned at its native 1-frame context to predict absolute states using gradient checkpointing. 

\noindent \textbf{SineNet}~\citep{zhang2024sinenet}: A multi-stage U-Net trained from scratch (2 waves, 4 layers/wave, 64 hidden channels). Uses zero padding to accommodate mixed open and periodic boundaries. The model is modified from the original paper to predict residuals instead of full field.

\noindent \textbf{FFNO}~\citep{tran2023factorized}: A factorized Fourier neural operator trained from scratch (16 modes, width 64, 4 layers) as a lightweight reference. The model is modified from the original paper to predict residuals instead of full field.

\noindent \textbf{Baselines (non-learned)}: \emph{Mean field} predicts a static velocity field derived from an exponential moving average of training frames. \emph{Self-advection} evolves the last frame via a semi-Lagrangian step that passively transports myosin alongside the flow:
  \begin{equation}
    u(x) \;\leftarrow\; u\!\left(x - \Delta t \, u(x)\right)
    \label{eq:selfadvection}
  \end{equation}

\begin{table}[t]
  \centering
  \caption{\textbf{Overview of Models and Baselines}. Context indicates input frames; Target denotes whether the model predicts the residual relative to the last frame or the absolute state.}
  \label{tab:models}
  \small
  \begin{tabular}{llccl}
    \toprule
    Model & Parameters & Context & Initialization & Target \\
    \midrule
    \textsc{Walrus}            & 1.3B          & 10 & Pretrained          & Residual \\
    \textsc{Walrus (scratch)}            & 1.3B          & 10 & Scratch          & Residual \\
    \textsc{Walrus}\,+\,CRPS     & 1.3B + AdaLN  & 10 & Finetuned \textsc{Walrus} & Residual \\
    Poseidon-L                   & 629M          & 1  & Pretrained          & Absolute \\
    SineNet                      & 68M           & 10 & Scratch             & Residual \\
    FFNO                         & 1.1M          & 10 & Scratch             & Residual \\
    Mean field                   & ---           & --- & ---                 & Absolute \\
    Self-advection               & ---           & 1  & ---                 & Absolute \\
    \bottomrule
  \end{tabular}
\end{table}

\paragraph{Checkpoint selection and Evaluation.}
Checkpoints are saved every 10 epochs. For each model, we select the checkpoint with the lowest validation VRMSE (averaged across rollout frames and validation embryos) and evaluate it unchanged on test, temperature, and mutant datasets for zero-shot testing. Following The Well benchmark~\citep{ohana2024well}, we report normalized VRMSE for cross-dataset comparability. Due to short trajectory lengths (50--90 frames), rollouts start immediately after the context window.

\section{Results}
We evaluate all models in-distribution on held-out wild-type (WT) embryos, and then zero-shot under two families of perturbation: (i) temperature shifts, which rescale the timescale of morphogenesis while preserving its spatial organization, and (ii) genetic mutations, which remove or distort flow modules. Evaluation metrics are defined in Appendix \ref{app:metrics}.

\subsection{Wild-type}
\begin{figure}[htbp]
    \centering
    \includegraphics[width=\textwidth]{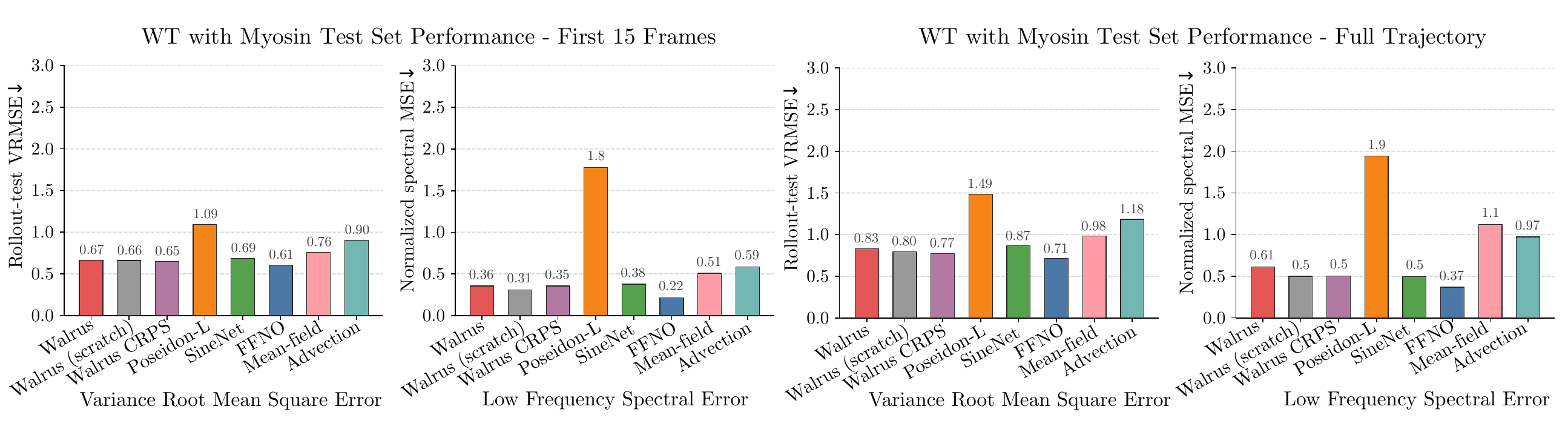}
    \caption{\textbf{In-distribution evaluation on the wild-type (WT) with-myosin test set.}
    Rollout-test VRMSE and low-frequency normalized spectral MSE (lower is better) for all
    models, evaluated over the first 15 rollout frames (left) and the full trajectory
    (right). FFNO attains the lowest error on both metrics at both horizons. Figure~\ref{fig:wt_test_app_results} reports the same evaluation for models trained
    without myosin.}
    \label{fig:wt_test_results}
\end{figure}

On the held-out WT embryos, FFNO is the strongest model on both rollout-test VRMSE and low-frequency normalized spectral MSE, at the 15-frame horizon and over the full trajectory alike (Figure~\ref{fig:wt_test_results}). The Walrus variants are competitive but do not lead, and the pretrained Poseidon-L is the weakest of the learned models. See Figure \ref{fig:wt_rollout} for an example of a predicted vs.\ true rollout for the Walrus CRPS model on a held-out test sample.

We view this gap as a characteristic of the evaluation regime rather than evidence against
pretraining. Because the WT test set shares the exact genotype and imaging temperature of the
training set, tissue flow follows a highly stereotyped spatiotemporal program across embryos, so
generalization requirements are minimal. A compact model trained from scratch (FFNO, 1.1M
parameters) can allocate its entire capacity to fitting the few dominant, low-wavenumber modes
that carry most of the signal. FFNO's margin is widest on the low-frequency spectral metric,
consistent with an advantage concentrated in these large-scale, repeatable components.
Conversely, the pretrained Walrus backbone retains broad representations spanning nineteen
distinct continuum systems, most of which are inactive in this context, and finetuning on only
12 embryos provides insufficient signal to fully specialize these features. Consistent with this,
a Walrus model trained from scratch matches or slightly outperforms its fine-tuned counterpart
in both settings (Figures~\ref{fig:wt_test_results} and~\ref{fig:wt_test_app_results}). In-distribution benchmarks inherently reward domain
specialization, whereas the true utility of general pretraining emerges under distribution shift.

\subsection{Zero-Shot}
After finetuning on the WT Sqh--mCherry embryos, we ran each model zero-shot on inputs perturbed by temperature or by genetic mutation \citep{lefebvre2026dynamicatlas}, with no additional training or adaptation. We ask how far the physics-pretrained representations transfer to biological perturbations never seen during finetuning. For the sake of space, we do not include advection and mean-field in the plots below.

\subsubsection{Temperature Varying}
\begin{figure}[t]
  \centering
  \includegraphics[width=\linewidth]{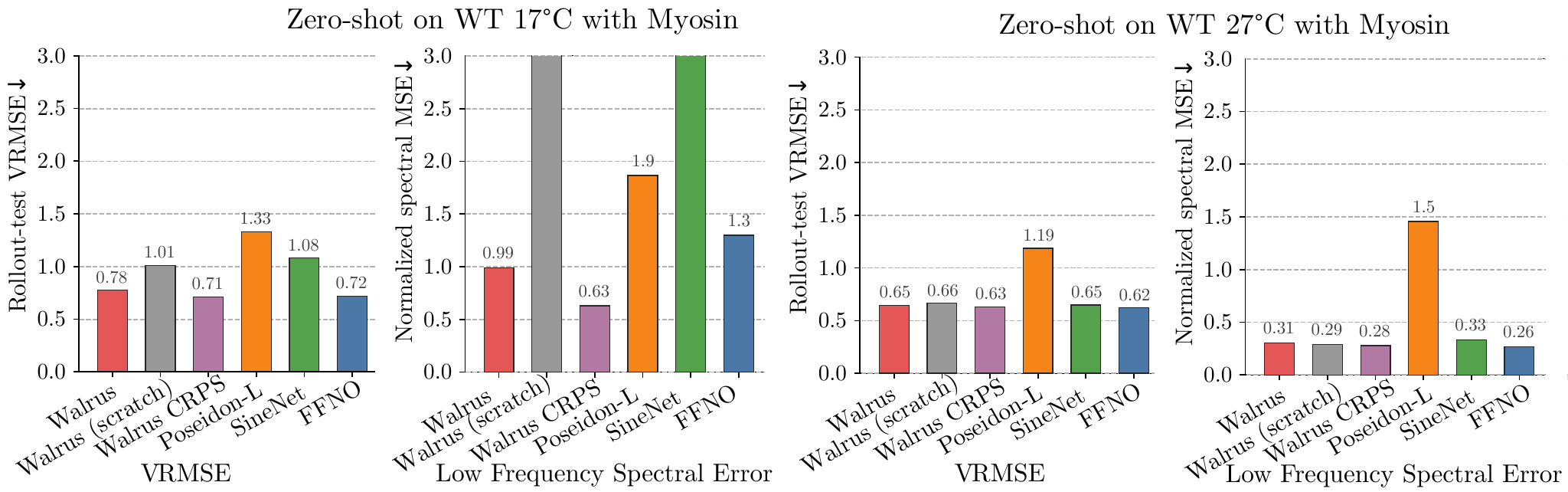}
  \caption{\textbf{Zero-shot performance evaluation on wild-type (WT) morphogenesis at
    $17^\circ\text{C}$ and $27^\circ\text{C}$.} Model accuracy is evaluated over the first 15 rollout
    frames using Variance Root Mean Square Error (VRMSE) and normalized spectral MSE restricted to low spatial frequencies (lower is better, showing learned models only). Bars exceeding the axis limit of $3.0$ are truncated. The
structure of the results differs by temperature: at
$27^\circ$C all models but Poseidon-L fall within a narrow band, so the ordering among them is not
meaningful, whereas at $17^\circ$C a clear split emerges between Walrus\,+\,CRPS, FFNO and Walrus
on one side and Walrus (scratch), SineNet and Poseidon-L on the other. See Figure~\ref{fig:zero_shot_eval_full} for the full rollout plot. See Figure~\ref{fig:zero_shot_eval_no_myosin} for models trained without myosin.}
  \label{fig:zero_shot_eval}
\end{figure}

WT training embryos were recorded at ambient ($\sim22^\circ$C) temperature. For zero-shot evaluation, we tested predictive performance on WT embryos recorded at $17^\circ$C ($n=2$) and $27^\circ$C ($n=3$). As reported by \citet{lefebvre2026dynamicatlas}, temperature monotonically alters flow speed without changing the spatial pattern of tissue motion. The perturbation therefore rescales the developmental timescale.

This is the first setting in which the in-distribution ranking breaks down, and it does so
asymmetrically (Figure~\ref{fig:zero_shot_eval}). At $27^\circ$C every model but Poseidon-L stays
near its in-distribution accuracy and FFNO still narrowly leads, by very small margins of $0.01$--$0.02$. At $17^\circ$C, errors rise sharply for all the baselines, and Walrus\,+\,CRPS becomes the best model on both metrics (VRMSE $0.71$, low-frequency spectral MSE $0.63$, against $0.72$ and $1.3$ for FFNO), while
Walrus trained from scratch and SineNet leave the plotted range entirely. Over the full trajectory
Walrus\,+\,CRPS is best or tied-best in all four panels (Figure~\ref{fig:zero_shot_eval_full}). A model that has fit the WT program tightly has also fit the \emph{rate} at which that program unfolds, so an embryo running faster or slower than any seen in training is out of distribution even though every spatial structure in it is familiar. Walrus is pretrained across continuum systems spanning a wide range of characteristic velocities and timescales, which makes its representation of the same spatial pattern less tied to one particular rate. 

\begin{figure}[htbp]
    \centering
    \includegraphics[width=\textwidth]{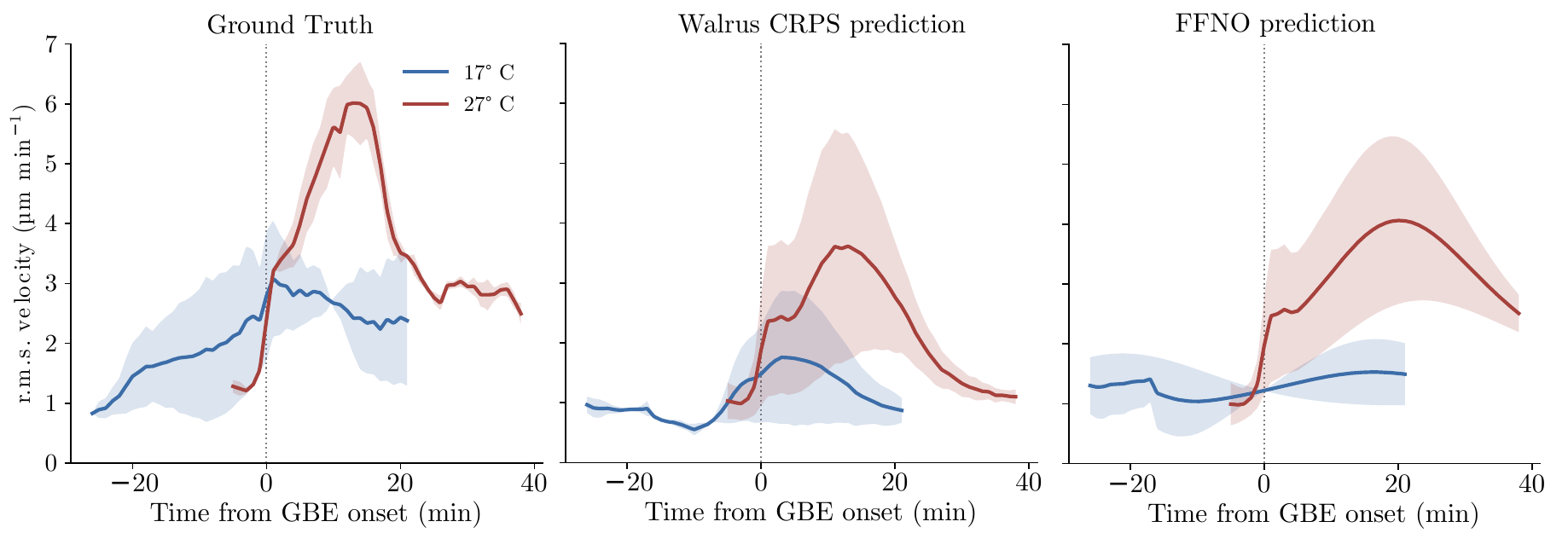}
    \caption{\textbf{Root-mean-square (RMS) surface velocity trajectories across thermal regimes.} For the two leading models of Figure~\ref{fig:zero_shot_eval} (Walrus\,+\,CRPS and FFNO), predicted
    embryo surface kinematics are compared against ground truth at $17^\circ\text{C}$ and
    $27^\circ\text{C}$, aligned to Germ Band Extension (GBE) onset. Shaded bands denote spread across embryos. Both models recover the ordering of the two temperatures and the post onset rise of the $27^\circ$C trajectory, but both underestimate peak RMS velocity by roughly $40\%$. Walrus\,+\,CRPS
    reproduces the $17^\circ$C peak near GBE onset at reduced amplitude, whereas FFNO predicts a nearly flat $17^\circ$C trajectory and places the $27^\circ$C peak $\approx7$ min late.}
    \label{fig:rms_velocity}
\end{figure}

Figure~\ref{fig:rms_velocity} shows that both leading models capture the general effect of temperature on kinematics without having seen either condition during training. However, they struggle with exact magnitudes, both underestimating peak RMS velocity by roughly 40\%. Furthermore, only Walrus\,+\,CRPS correctly captures the $17^\circ$C peak near GBE onset. FFNO predicts a nearly flat trajectory and delays the $27^\circ$C peak by $\approx7$ min. Further confirming this adaptation, Figure~\ref{fig:hist_17_deg_myosin} demonstrates that the Walrus CRPS model does not merely memorize WT predictions, but captures the distinctly narrower velocity distribution shape of the $17^\circ$C zero-shot data.

\subsubsection{Genotype Varying}
To probe a stronger form of distribution shift, we benchmarked zero-shot transfer across three genetic mutations without myosin inputs. The mutants are described below and span a spectrum of increasing morphological disruption relative to the wild-type (WT) training baseline.

\noindent \textbf{Mild Mutation: \textit{eve[r13]}.} (n = 9) Tissue flow speed within the germ band module is reduced, though the spatial flow pattern remains largely preserved.

\noindent \textbf{Moderate Mutation: \textit{twi[ey53]}.} (n = 6) Dorsal contraction and ventral furrow modules are completely absent, while germ band flow patterns are maintained.

\noindent \textbf{Severe Mutation: \textit{spz} / \textit{spaetzle}.} (n = 4) All major movement modules are absent, resulting in a near-total collapse of tissue flow.

Unlike the temperature shift, these perturbations change which developmental modules are present
at all, and error grows accordingly. For every model except Poseidon-L, rollout-test VRMSE and
low-frequency normalized spectral MSE over the first 15 frames increase monotonically with
severity (Figure~\ref{fig:mutation_zero_shot}). The reordering seen under temperature becomes
unambiguous for the two stronger perturbations: on \textit{twi[ey53]} and \textit{spz},
Walrus\,+\,CRPS attains the lowest error on both metrics (VRMSE $0.99$ and $1.11$; spectral MSE
$0.86$ and $1.7$), and its margin over FFNO widens with severity (VRMSE gap $0.07$, $0.23$, $0.36$
across the three mutants). On the mildest mutant, \textit{eve[r13]}, Walrus\,+\,CRPS is not the
leader, however all models except Poseidon-L and FFNO fall within $0.08$ of one another on both metrics.

\paragraph{Takeaways.} Together, these three evaluations clarify a distinction that is easy to conflate. For predictable, in-distribution wild-type dynamics, a small model trained directly on the target behavior performs best, while physics pretraining offers little value (Fig.~\ref{fig:wt_test_results}). However, under target distribution shifts, first in timescale, then in the presence of specific flow modules, that ordering progressively reverses, to the extent that for the two stronger mutants the pretrained continuum-dynamics representation is decisive. Notably, this holds under a demanding rollout protocol: Walrus conditions on just ten ground-truth frames before predicting the remainder autoregressively, meaning its context contains no real observations after step ten. The perturbation-specific cues it relies on must therefore be present in those first ten frames. This is well before the obvious morphological divergences that define each mutant phenotype appear. That subtle early variations in the velocity field are enough to forecast distinct trajectory paths suggests these recordings carry far more usable early signal than visible morphology alone reveals. Identifying what that signal is remains an exciting next step.

\section{Conclusion and Discussion}
Here we present what is, to the best of our knowledge, a first attempt to study the transfer of Continuum Dynamics foundation models to the prediction of developmental dynamics in videos of the \textit{Drosophila} blastoderm. We finetuned Walrus on a small number of wild-type videos from the DynamicAtlas on the task of autoregressive next-frame prediction of tissue velocity fields. Finetuning on this type of data constitutes a substantial distribution shift with respect to the pretraining dataset: the embryo videos are experimental, while pretraining data is simulated. Moreover, Walrus has not been trained on the particular combination of physical processes that drive embryonic development, although several of these physics are represented individually in its pretraining dataset.

The finetuned model is able to capture the trends of several qualitative features with biological relevance, such as changes in flow magnitude with temperature and mean velocity evolution that characterize large-scale tissue-level dynamics. In general, finetuned Walrus performs competitively with the other baseline models on wild-type evaluations, while achieving the best performance in out-of-distribution zero-shot experiments on the mutations \textit{twi[ey53]} and \textit{spz}, despite being trained only on wild-type data. Mutation data are both scarcer and more diverse than wild-type data, and therefore represent a regime in which leveraging representations learned from diverse physical systems could be particularly useful. Finetuned Walrus is also the best or nearly the best performer when tested on unseen embryo videos recorded at 17$^\circ$C and 27$^\circ$C, with training data only containing samples at room temperature 22$^\circ$C. Temperature affects the rate of development but does not introduce phenotypical changes; models trained from scratch only on wild-type room temperature may struggle to generalize in comparison to models trained on a wide set of PDEs and configurations. Further study will also investigate the role of CRPS in the results and understand the performance asymmetry with respect to the temperature ranges considered.

Overall, our results suggest that representations learned from simulated systems spanning a diverse set of continuum dynamics may improve prediction of morphogenetic dynamics in systems that admit a continuum, fluid-like description, particularly in low-data regimes such as the mutant conditions considered here. 
The conclusions of this study are preliminary and limited to a particular organism, developmental stage, and set of conditions that are favorable to a Continuum Dynamics description. Another limitation is the scarcity of biological baselines that can be compared directly with the models considered here. A further limitation, particularly relevant for domain experts, is that successful next-frame prediction does not necessarily translate to an understanding of the physical mechanisms driving the observed dynamics. It therefore remains unclear to what extent the internal representations of these foundation models correspond to physically and biologically meaningful concepts.

This work motivates the further study of morphogenesis as a downstream application of physics foundation models. Immediate next steps include a more extensive study of zero-shot performance across mutations with a range of phenotypes, as well as restricting predictions to localized developmental regions and extending the framework to other species (e.g., zebrafish \citep{kimmel1995stages}) or \textit{in vitro} organoids. Studying the data efficiency of foundation models could also guide experimental design by determining the optimal temporal and spatial resolutions required for accurate forecasting. In a similar line, varying the rollout starting time and context window could help to quantify when development becomes predictable. An exciting direction could be to use the later studies to identify if there are early features in the videos, hidden to the human eye, that contribute to predictability.

Furthermore, beyond global aggregate metrics like VRMSE, future evaluations should isolate critical spatial structures, such as attractors, repellers, or Lagrangian Coherent Structures \citep{serra2020dynamic}. This would test the model's ability to predict major morphogenetic features, like the precise onset of Germ Band Extension or the absence of the ventral furrow. Lastly, an important long-term question is whether we can use finetuned representations to understand the physical drivers of wild-type zero-shot generalization; future work should trace which specific pretraining physics PDEs contribute the most to performance, revealing the coarse continuum dynamics that tissue flow most closely resembles across space and time.

\section*{Acknowledgments}
We thank Daniel Alber and Hayden Nunley for helpful discussions and valuable feedback. We would like to acknowledge the support of the Simons Foundation and of Schmidt Sciences. This work was supported in part by the AI2050 program at Schmidt Sciences (Grant G-25-70028). Dr. Mukhopadhyay thanks the Infosys-Cambridge AI center for support.  Louisa Cornelis acknowledges support from the NSF Graduate Research Fellowship Program (NSF GRFP) Grant 2139319. 

\bibliographystyle{plainnat}
\bibliography{references}

\appendix
\section{Extended Data Processing Methods}
\label{app:extended_methods}

The following sections describe the methods for velocity field generation and quality assessment, as well as the procedures used for temporal alignment of recordings across the dataset.

\subsection{Velocity Field Generation and Quality Control}
\label{app:qc}
PIV was run on the atlas pullbacks after an image downscaling factor of 0.4. The retained run utilized \texttt{EdgeLength = 15}, corresponding to four interrogation passes ($120 \to 60 \to 30 \to 15$ pixels), a final interrogation window of $9.8\,\mu\mathrm{m}$, and vector spacing of $5.2\,\mu\mathrm{m}$. Grid cells were defined as measured only if raw PIV components were finite and filtered components were unchanged. The retained cohort median for interpolated vectors was $9.6\%$.

\begin{figure}[htbp]
    \centering
    \includegraphics[width=\linewidth]{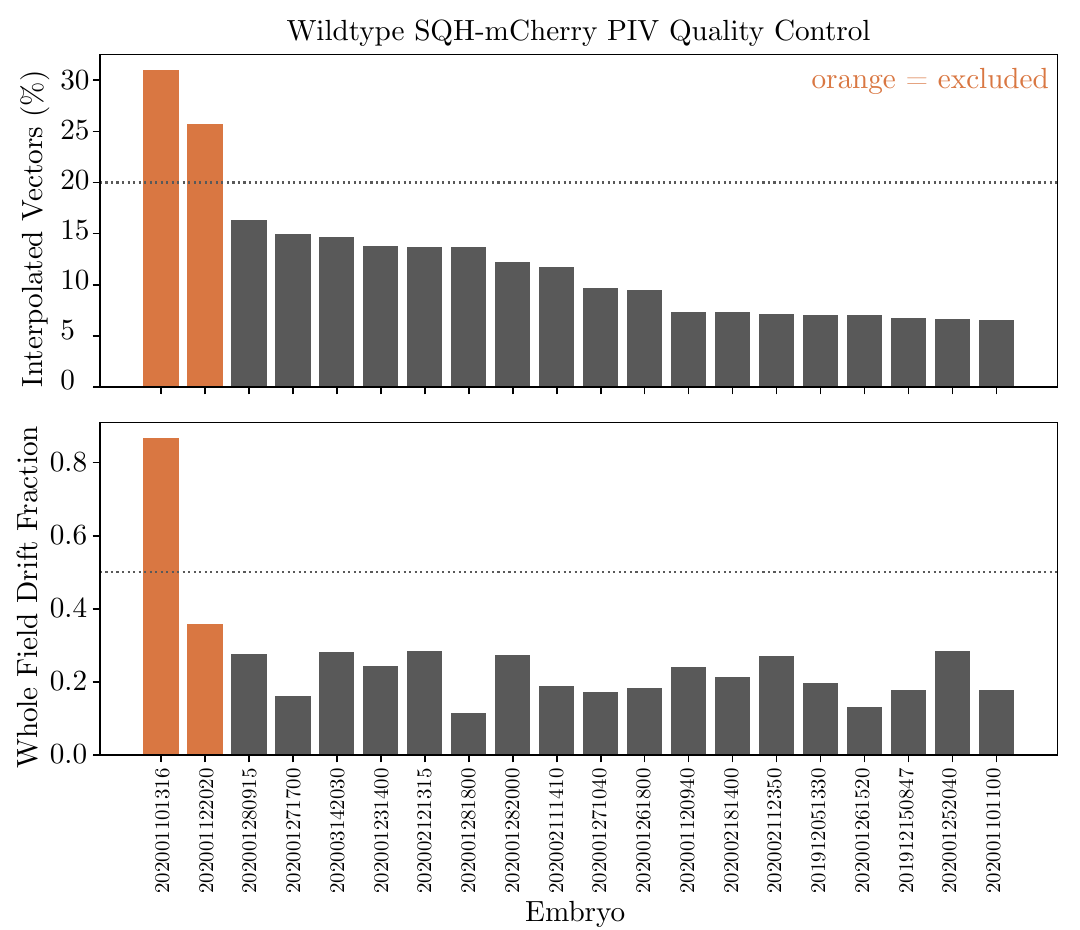}
    \caption{\textbf{Velocity-field quality control for WT Sqh--mCherry recordings.} The retained cohort is defined by field-intrinsic PIV quality and whole-field drift, not by image-warp residuals for the deep maximum-projection recordings.}
    \label{fig:velocity-qc}
\end{figure}

Two recordings were excluded based on field-intrinsic quality: \texttt{202001101316} (unwrapping artifact, $86.6\%$ whole-field translation) and \texttt{202001122020} (luminosity fluctuation). Four deep maximum projection recordings (\texttt{201912051330}, \texttt{201912150847}, \texttt{202001101100}, \texttt{202001120940}) were retained as they demonstrated standard field-intrinsic QC (Fig.~\ref{fig:velocity-qc}).

\subsection{Temporal Alignment Details}
\label{app:temporal}
\begin{figure}[htbp]
    \centering
    \includegraphics[width=\linewidth]{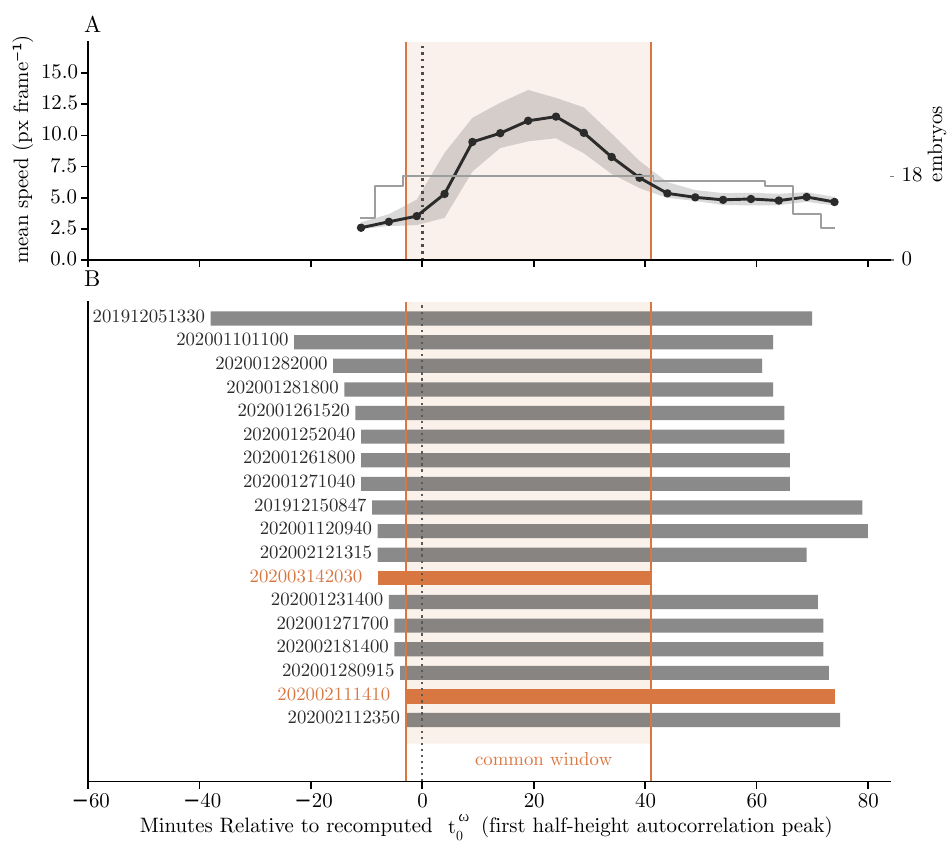}
    \caption{\textbf{Temporal alignment.} Retained WT Sqh--mCherry recordings relative to $t_0^\omega$.}
    \label{fig:temporal-alignment}
\end{figure}

We measure the similarity in tissue rotation $\omega=\nabla\times\mathbf{v}$ between frames using vorticity correlation $M(t_i,t_j)$ \citep{lefebvre2026dynamicatlas}:
\begin{equation}
  M(t_i,t_j)
  =\frac{\displaystyle\int\!\big(\omega_i-\bar\omega_i\big)
                  \big(\omega_j-\bar\omega_j\big)\,dA}
        {\displaystyle\Big(\int\!\big(\omega_i-\bar\omega_i\big)^2\,dA\Big)^{1/2}
         \Big(\int\!\big(\omega_j-\bar\omega_j\big)^2\,dA\Big)^{1/2}} .
\end{equation}
The rate of change along the first time axis is scored as $s(t)=\sum_{t'}\partial_t M(t,t')$. We defined the kinematic landmark $t_0^\omega$ as the first local maximum of $s(t)$ reaching at least half its global maximum:
\begin{equation}
  t_0^\omega=\min\{t:\ t\ {\rm is\ a\ local\ maximum\ of}\ s,\quad
  s(t)\geq 0.5\max_{t'}s(t')\}.
\end{equation}
This high-resolution robustness rule reduced the mean absolute difference from the atlas landmark $t_0^V$ from 12.44 to 3.94 minutes (Table \ref{tab:t0}; Fig.~\ref{fig:temporal-alignment}).

\begin{table}[htbp]
    \centering
    \small
    \begin{tabular}{lccccccc}
        \toprule
        & & & atlas & \multicolumn{2}{c}{released DynamicAtlas PIV} & \multicolumn{2}{c}{our video} \\
        \cmidrule(lr){5-6}\cmidrule(lr){7-8}
        recording & split & $T$ frames & $t_0^V$ & $t_0^\omega$ & $\Delta$ & $t_0^\omega$ & $\Delta$ \\
        \midrule
        \texttt{201912051330} & train & 109 & 41 & -- & -- & 39 & $-2$ \\
        \texttt{201912150847} & test & 89 & 10 & 6 & $-4$ & 10 & $0$ \\
        \texttt{202001101100} & valid & 87 & 18 & 16 & $-2$ & 24 & $+6$ \\
        \texttt{202001120940} & train & 89 & 8 & 9 & $+1$ & 9 & $+1$ \\
        \texttt{202001231400} & valid & 78 & 8 & 7 & $-1$ & 7 & $-1$ \\
        \texttt{202001252040} & train & 77 & 14 & -- & -- & 12 & $-2$ \\
        \texttt{202001261520} & train & 78 & 15 & -- & -- & 13 & $-2$ \\
        \texttt{202001261800} & train & 78 & 12 & 11 & $-1$ & 12 & $0$ \\
        \texttt{202001271040} & test & 78 & 15 & 15 & $0$ & 12 & $-3$ \\
        \texttt{202001271700} & train & 78 & 17 & -- & -- & 6 & $-11$ \\
        \texttt{202001280915} & train & 78 & 23 & 23 & $0$ & 5 & $-18$ \\
        \texttt{202001281800} & train & 78 & 6 & 6 & $0$ & 15 & $+9$ \\
        \texttt{202001282000} & train & 78 & 20 & -- & -- & 17 & $-3$ \\
        \texttt{202002111410} & test & 78 & 11 & -- & -- & 4 & $-7$ \\
        \texttt{202002112350} & train & 79 & 5 & -- & -- & 4 & $-1$ \\
        \texttt{202002121315} & valid & 78 & 7 & -- & -- & 9 & $+2$ \\
        \texttt{202002181400} & train & 78 & 8 & -- & -- & 6 & $-2$ \\
        \texttt{202003142030} & train & 50 & 10 & 9 & $-1$ & 9 & $-1$ \\
        \midrule
        mean $|\Delta|$ (median) & & & & \multicolumn{2}{c}{$1.11$ ($-1$)} & \multicolumn{2}{c}{$3.94$ ($-1.5$)} \\
        \bottomrule
    \end{tabular}
    \caption{\textbf{Recomputed vorticity-autocorrelation.} Landmark $t_0^\omega$ against the atlas landmark $t_0^V$. The $T$ column gives the number of velocity frames available for each Well-formatted recording.}
    \label{tab:t0}
\end{table}

\section{Evaluation Metrics}
\label{app:metrics}

Unless noted, metrics are computed in physical units on open-loop rollouts
after the initial context window. For the stochastic \textsc{Walrus}+CRPS
model we evaluate the ensemble mean. Scores are averaged over space, then over
time and fields, then over embryos, then equally over datasets, matching
The Well benchmark~\cite{ohana2024well}.

\subsection{Variance-scaled RMSE (VRMSE)}

VRMSE is a spatially averaged RMSE scaled by the spatial standard deviation of
the ground truth, so a constant-mean prediction scores~$\approx 1$ and a
perfect prediction scores~$0$:
\begin{equation}
  \mathrm{VRMSE}(\hat{u},u)
  =
  \sqrt{
    \frac{
      \langle (\hat{u}-u)^{2}\rangle_{\mathbf{x}}
    }{
      \mathrm{Var}_{\mathbf{x}}(u)+\varepsilon
    }
  },
\end{equation}
with $\varepsilon=10^{-5}$. We report both the full rollout and a short
horizon (first 15 predicted frames).

\subsection{Binned spectral error}

To separate large-scale structure from small-scale detail we use The Well's \citep{ohana2024well}
binned spectral MSE. Let $\widehat{\Delta}(\mathbf{k})$ be the spatial Fourier
transform of $\hat{u}-u$. Isotropic wavenumber
$k=\lvert\mathbf{k}\rvert$ is partitioned into three approximately
log-spaced bins from $k=0$ to the Nyquist scale (low, medium, and high
frequency). The error in bin~$b$ is the mean squared residual power in that
shell. We report the \emph{normalized} form
\begin{equation}
  \mathrm{NMSE}_{b}
  =
  \frac{
    \sum_{\mathbf{k}\in b}\lvert\widehat{\Delta}(\mathbf{k})\rvert^{2}
  }{
    \sum_{\mathbf{k}\in b}\lvert\hat{u}_{\mathrm{GT}}(\mathbf{k})\rvert^{2}
    +\varepsilon
  },
\end{equation}
so each band is relative to the ground-truth energy in that band. Low
frequency corresponds to large-scale, smooth structure; high frequency
corresponds to small-scale detail. Higher frequencies are omitted from spectral evaluation because system complexity and noise cause all models to perform poorly ($\text{error} \ge 1.0$) at mid and high frequencies. 

\subsection{Myosin Tensor Validation}
\label{app:myosin}
Testing a literature-style cytosolic normalization yielded good signed orientation agreement (0.917) but poor component correlations (mean $M_{xx}$ correlation $-0.097$). Amplitudes were lower by roughly a factor of four. Running the windowed-Radon transform directly on raw maximum-projection intensity without floor subtraction restored full-frame agreement with provided atlas references (mean $M_{xx}$ correlation 0.706, norm correlation 0.694, signed orientation agreement 0.988). We therefore utilize the \texttt{raw\_max\_intensity} tensor to best preserve the concentration-weighted spatial field.

\section{Wild-type Training Supplementary Data}

Additional plots for the in-distribution model evaluation, showing rollout visualizations and model performance. 
\begin{figure}[!htbp]
    \centering
    \includegraphics[width=0.8\textwidth]{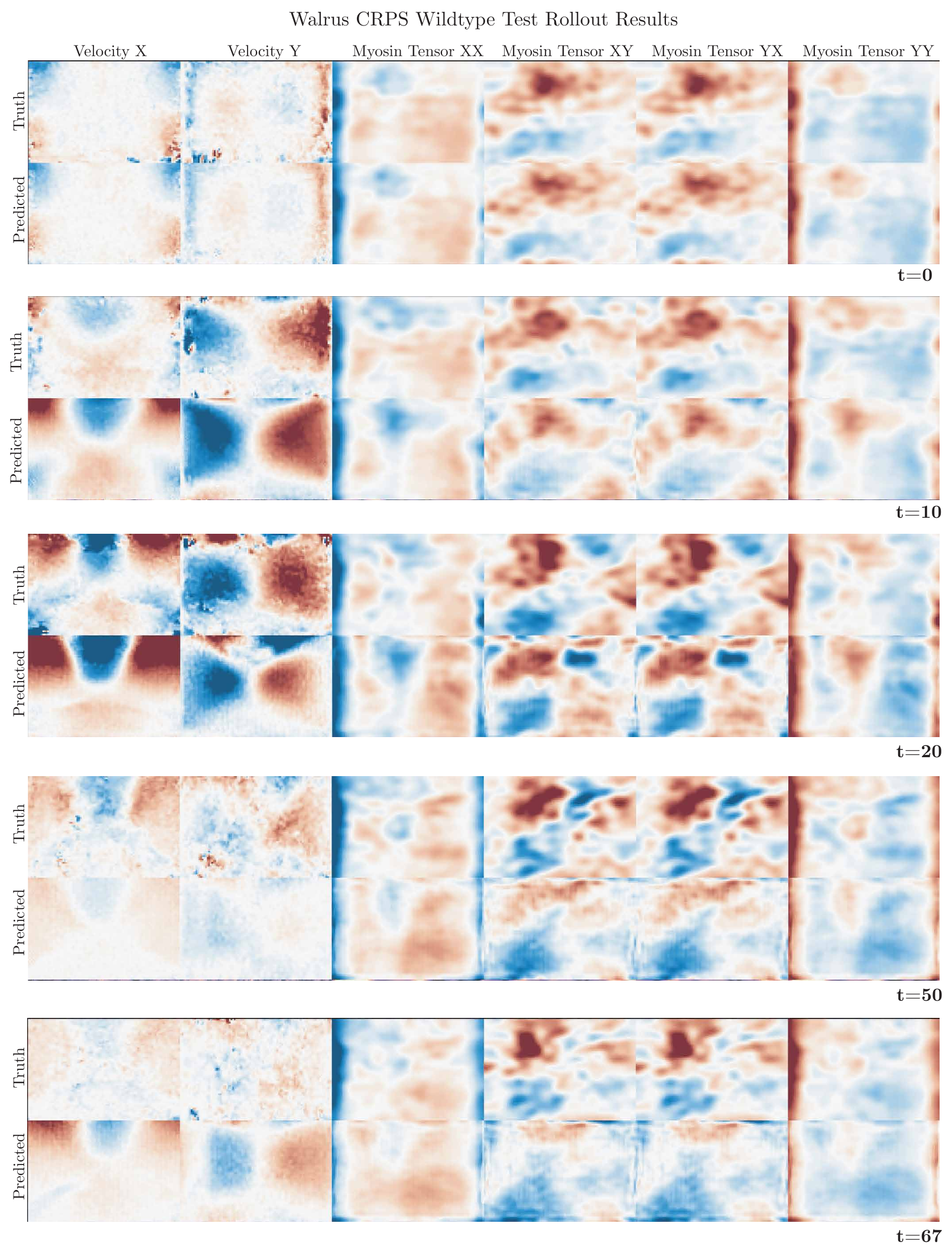}
    \caption{\textbf{Rollout results for Walrus CRPS.} Model trained on wild-type embryos, shown on a held-out wild-type test embryo at frames 0, 10, 20, 50 and 67.}
    \label{fig:wt_rollout}
\end{figure}

\begin{figure}[!htbp]
    \centering
    \includegraphics[width=\textwidth]{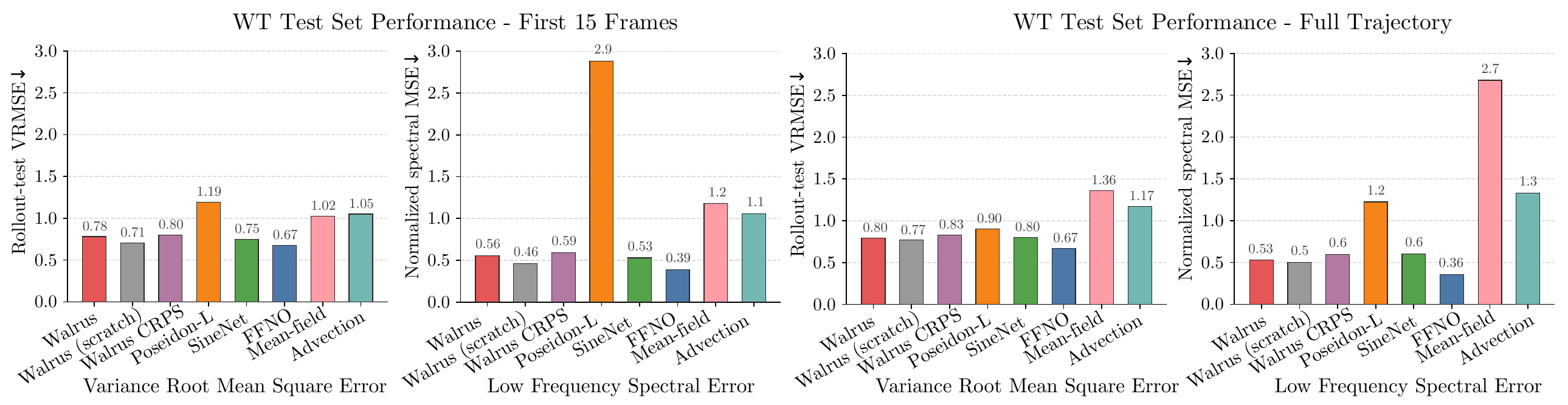}
    \caption{\textbf{In-distribution evaluation on the wild-type (WT) test set without myosin.} Performance across models measured by Rollout-test VRMSE and low-frequency normalized spectral MSE evaluated over the first 15 rollout frames (left) and the full trajectory (right). Across both time horizons and metrics, FFNO consistently emerges as the top-performing model, achieving the lowest overall errors.}
    \label{fig:wt_test_app_results}
\end{figure}

\section{Temperature Zero-Shot Supplementary Data}
Additional plots for the zero-shot temperature varying model evaluation. 
\begin{figure}[!htbp]
  \centering
  \includegraphics[width=\linewidth]{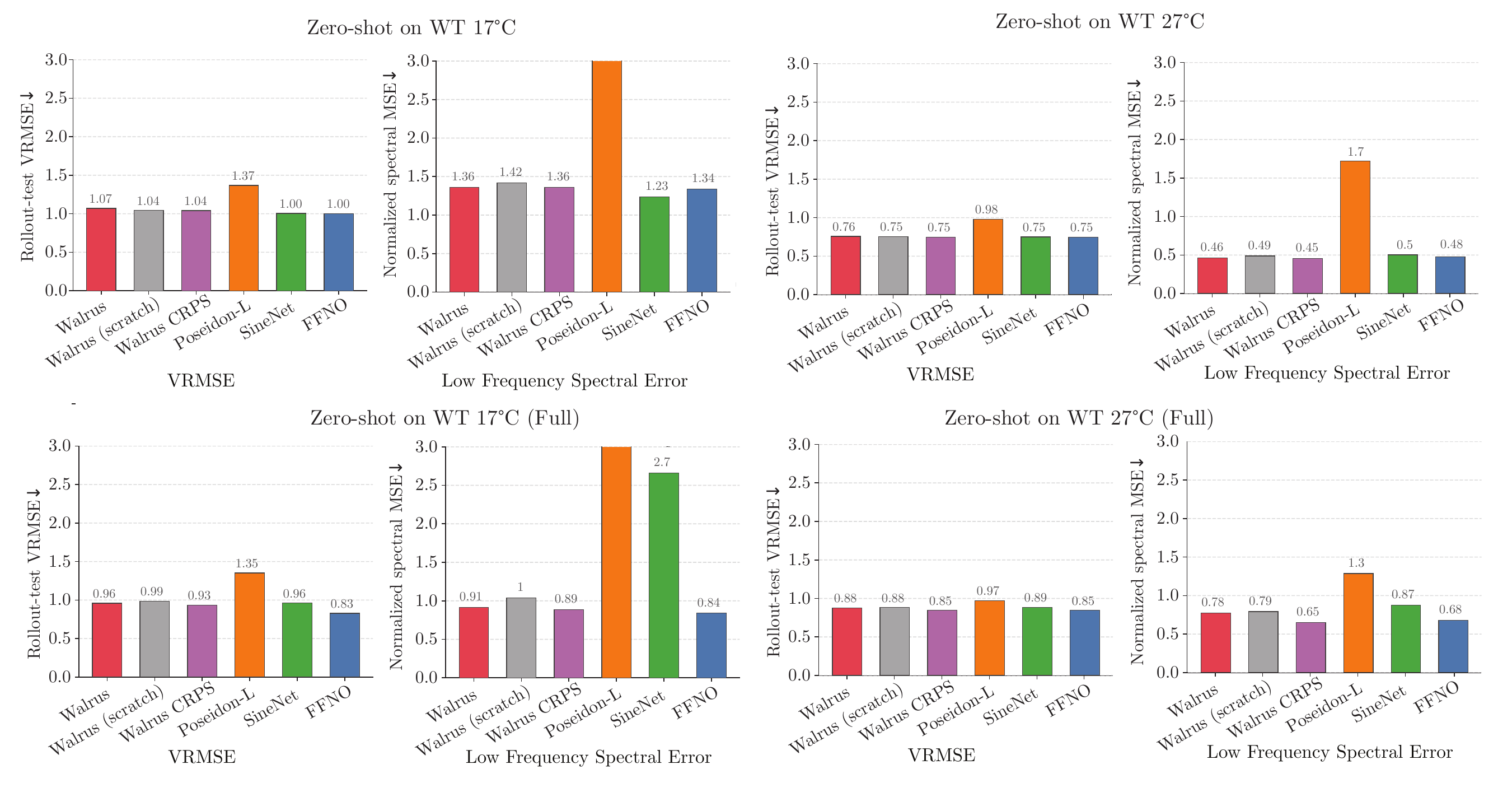}
  \caption{\textbf{Zero-shot performance evaluation, models trained without myosin.} Wild-type (WT) morphogenesis at $17^\circ\text{C}$ and $27^\circ\text{C}$ over the first 15 frames (top row) and the full rollout (bottom row). Bars exceeding the axis limit of $3.0$ are truncated. Walrus\,+\,CRPS leads at $27^\circ$C over the full trajectory and has a narrower gap at $17^\circ$C than during in-distribution.}
  \label{fig:zero_shot_eval_no_myosin}
\end{figure}

\begin{figure}[!htbp]
  \centering
  \includegraphics[width=\linewidth]{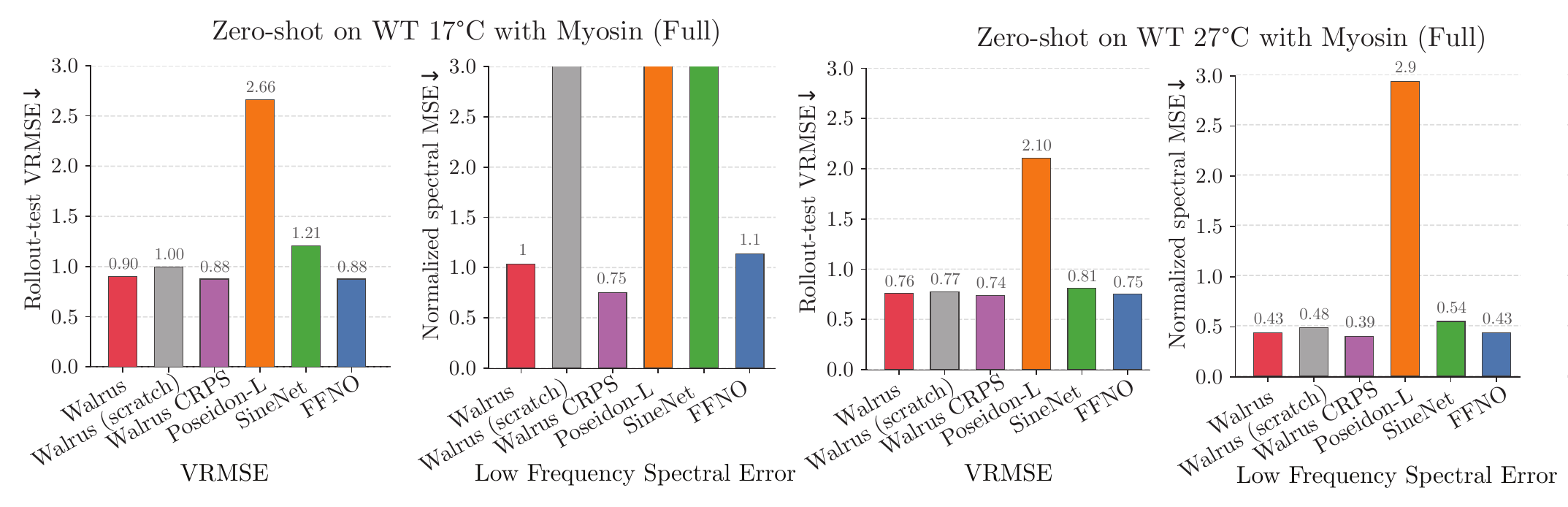}
  \caption{\textbf{Zero-shot performance evaluation over the full rollout trajectory, models trained
with myosin.} Wild-type (WT) morphogenesis at $17^\circ\text{C}$ and $27^\circ\text{C}$. Bars
exceeding the axis limit of $3.0$ are truncated. Walrus\,+\,CRPS is the best or tied-best model in
all four panels.}
  \label{fig:zero_shot_eval_full}
\end{figure}

\begin{figure}[!htbp]
    \centering
    \includegraphics[width=\textwidth]{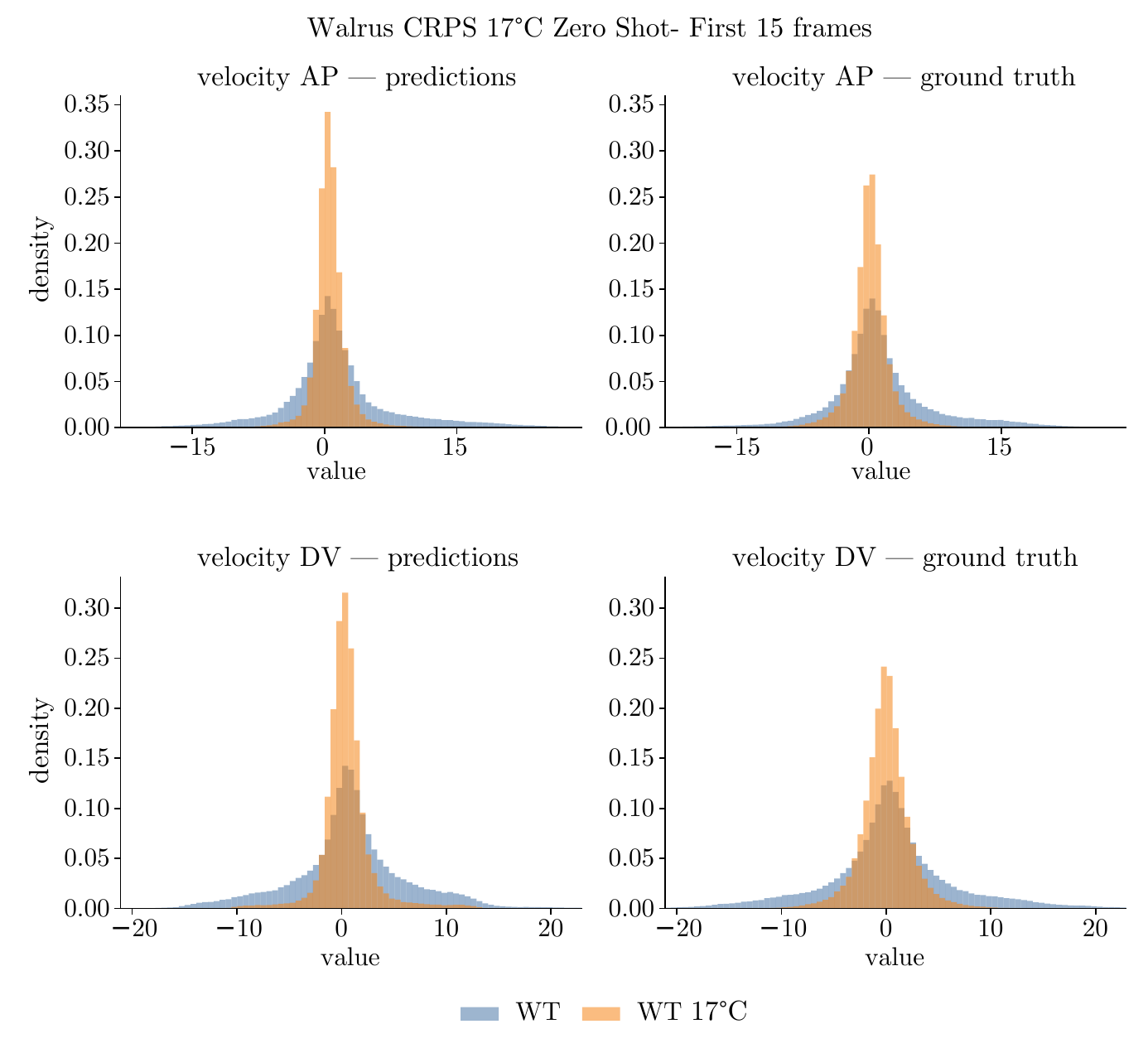}
    \caption{\textbf{Zero-Shot Velocity Distribution Adaptation.} Density histograms of physical-unit $v_{AP}$ (top) and $v_{DV}$ (bottom) velocities pooled over the first 15 predicted autoregressive frames. The original WT test set (blue) is compared against the zero-shot $17^\circ\text{C}$ dataset (orange). The left column displays the model's predictions, while the right displays the matching ground truth. The ground truth panels establish that the $17^\circ\text{C}$ dataset has a narrower velocity distribution than the WT data. Crucially, the prediction panels demonstrate that the model successfully adapts to this zero-shot distribution shift, accurately matching the narrower histogram shape of the $17^\circ\text{C}$ data rather than regurgitating the broader WT predictions. This plot is for the model trained with myosin.}
    \label{fig:hist_17_deg_myosin}
\end{figure}


\section{Mutation Zero-Shot Supplementary Data}

Plots for the zero-shot genotype varying model evaluation.

\begin{figure}[htbp]
    \centering
    \includegraphics[width=\textwidth]{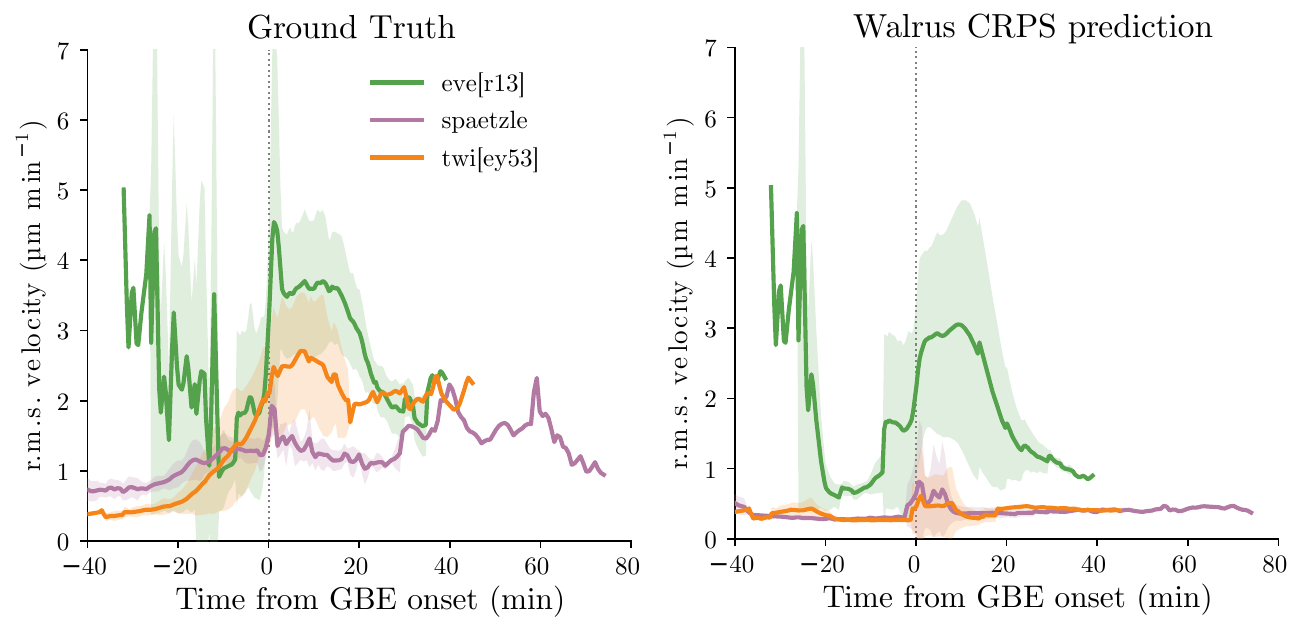}
    \caption{\textbf{RMS surface velocity trajectories across mutant phenotypes, Walrus CRPS.} Ground truth (left) and predictions (right) for \textit{eve[r13]}, \textit{twi[ey53]}, and \textit{spaetzle}, aligned to GBE onset. Shaded bands indicate the spread across embryos. For the mild \textit{eve[r13]} mutant, the model tracks the general rise and decay, though it underestimates the peak amplitude ($\approx3.1$ vs.\ $\approx4.5\,\mu$m\,min$^{-1}$). However, for the moderate \textit{twi[ey53]} and severe \textit{spaetzle} phenotypes, where morphogenetic movements are largely suppressed, the model struggles to meaningfully distinguish between their respective low-magnitude velocity trajectories.}
    \label{fig:rms_velocity_mutations}
\end{figure}

\begin{figure}[!htbp]
    \centering
    \includegraphics[width=0.9\textwidth]{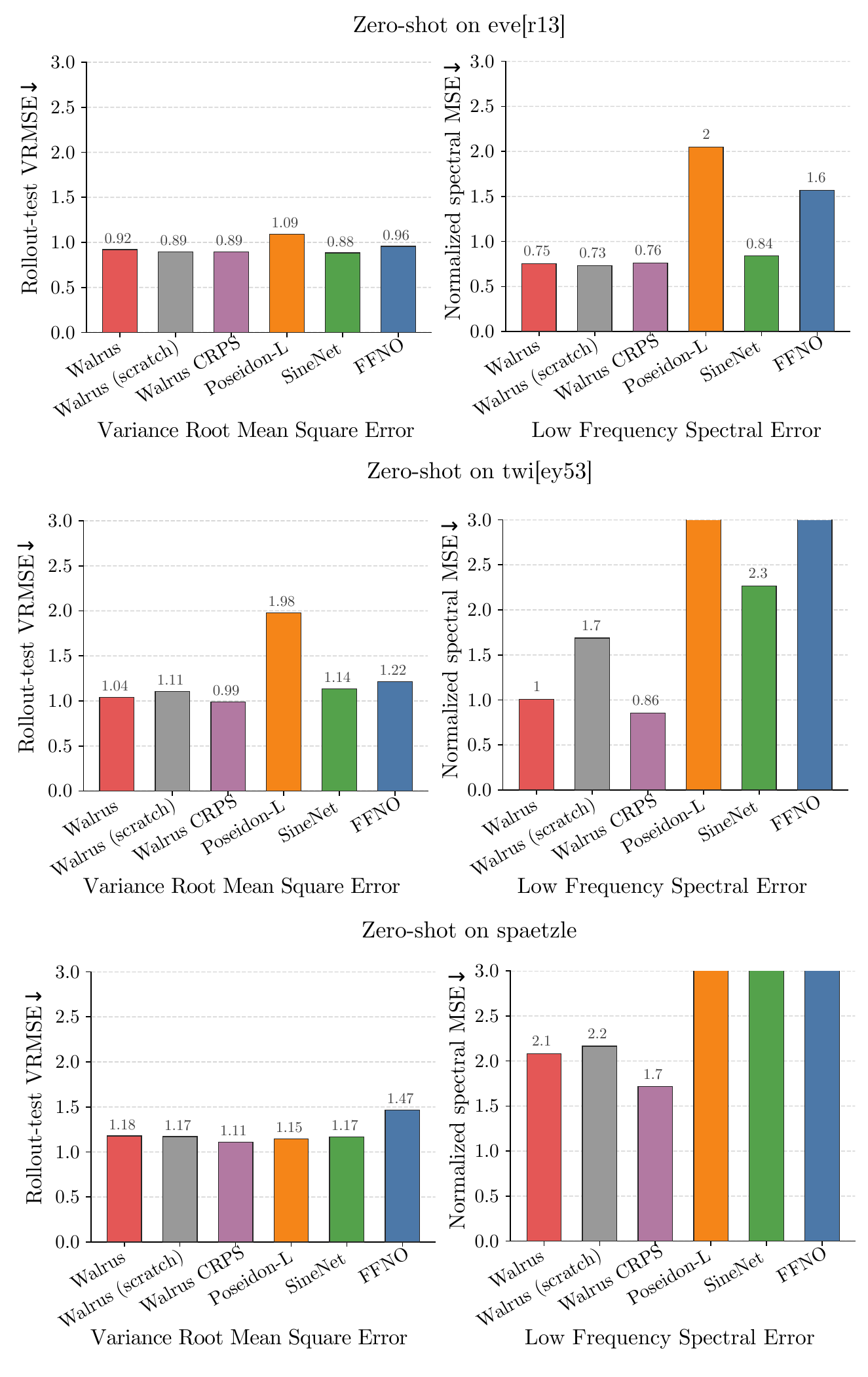}
    \caption{\textbf{Zero-shot generalization across genetic mutations of increasing severity (without myosin).} Model performance on the first 15 rollout frames for three mutant strains (one per row): \textit{eve[r13]} (mild speed reduction), \textit{twi[ey53]} (loss of dorsal contraction and ventral furrow modules), and \textit{spz} (severe loss of all developmental modules and flow). VRMSE (left) and low-frequency normalized spectral MSE (right) increase with severity for every model except Poseidon-L, whose VRMSE decreases on \textit{spz}. Bars exceeding the axis limit of $3.0$ are truncated. Walrus CRPS attains the lowest error on both metrics for \textit{spz} and \textit{twi[ey53]} and is within 0.01 of SineNet for \textit{eve[r13]}. On \textit{eve[r13]} SineNet achieves lowest on VRMSE and Walrus (scratch) achieves lowest on spectral MSE.}
    \label{fig:mutation_zero_shot}
\end{figure}

\clearpage

\end{document}